\documentclass[manuscript]{acmart}

\usepackage{sidecap}

\newcommand{\One}{{R1}}
\newcommand{\Two}{{R2}}
\newcommand{\Three}{{R3}}
\newcommand{\Four}{{R4}}
\newcommand{\Five}{{R5}}
\newcommand{\Six}{R6}
\newcommand{\Seven}{R7}
\newcommand{\Eight}{R8}
\newcommand{\Nine}{R9}
\newcommand{\Ten}{R10}
\newcommand{\Eleven}{R11}
\newcommand{\Twelve}{R12}
\newcommand{\Thirteen}{R13}
\newcommand{\Fourteen}{R14}
\newcommand{\Fifteen}{R15}
\newcommand{\Sixteen}{R16}
\newcommand{\Seventeen}{R17}

\usepackage{xcolor}

\copyrightyear{2021}
\acmYear{2021}
\setcopyright{rightsretained}
\acmConference[AIES '21]{Proceedings of the 2021 AAAI/ACM Conference
on AI, Ethics, and Society}{May 19--21, 2021}{Virtual Event, USA}
\acmBooktitle{Proceedings of the 2021 AAAI/ACM Conference on AI,
Ethics, and Society (AIES '21), May 19--21, 2021, Virtual Event,
USA}\acmDOI{10.1145/3461702.3462527}
\acmISBN{978-1-4503-8473-5/21/05}

\begin{document}
\fancyhead{}

\title{Machine Learning Practices Outside Big Tech:\\How Resource Constraints Challenge Responsible Development}


\author{Aspen Hopkins}
\authornote{Both authors contributed equally to this research.}
\affiliation{%
  \institution{Massachusetts Institute of Technology}
  \city{Cambridge}
  \country{USA}}
\email{dataspen@mit.edu}

\author{Serena Booth}
\authornotemark[1]
\affiliation{%
  \institution{Massachusetts Institute of Technology}
  \city{Cambridge}
  \country{USA}}
\email{sbooth@mit.edu}

\renewcommand{\shortauthors}{Hopkins and Booth}

\begin{abstract}

Practitioners from diverse occupations and backgrounds are increasingly using machine learning (ML) methods. Nonetheless, studies on ML Practitioners typically draw populations from Big Tech and academia, as researchers have easier access to these communities. Through this selection bias, past research often excludes the broader, lesser-resourced ML community---for example, practitioners working at startups, at non-tech companies, 
and in the public sector. These practitioners share many of the same ML development difficulties and ethical conundrums as their Big Tech counterparts; however, their experiences are subject to additional under-studied challenges stemming from deploying ML with limited resources, increased existential risk, and absent access to in-house research teams.
We contribute a qualitative analysis of 17 interviews with stakeholders from  organizations which are less represented in prior studies. We uncover a number of tensions which are introduced or exacerbated by these organizations' resource constraints---tensions between privacy and ubiquity, resource management and performance optimization, and access and monopolization. Increased academic focus on these practitioners can facilitate a more holistic understanding of ML limitations, 
and so is useful for prescribing a research agenda to facilitate responsible ML development for all.

\end{abstract}



\maketitle

\section{Introduction}

In past research analyzing ML practice, the vast majority of studies draw participants from Big Tech companies or academia~\cite{8046093, kim2016emerging, gamut2019, trustcorporatedata, 10.1145/3359313, zhang-2020-datascience,kaur2020interpreting,rakova2020responsible,holstein2019improving,beutel2019putting,amershi2019guidelines, muller2021designing,kim2016emerging,muller2019data}, with few exceptions \cite{pereira2020understanding,brennen2020people,hong2020human}. However, wealthy Big Tech and academic communities offer privileges and perspectives that are \emph{not} universally representative. For example, \citet{wired} chronicled how a collaboration between Google and Carnegie Mellon University collected 300 million labeled images and used fifty GPUs for two months---a scale of development which is increasingly the norm, yet is untenable for less resourced or less experienced organizations. This leads to the question: how well do past studies of Big Tech and academic practitioners encompass the needs of other data and ML workers?

\citet{pereira2020understanding} observed that the diversity of data science teams' compositions, goals, and processes remains understudied---particularly for practitioners outside of Big Tech. This is certainly not the only understudied component of data and ML work outside of Big Tech and academia. We ask: what problems do smaller companies, organizations, and agencies face? What are their practices? How can the AI research community ensure our work is targeted not just at well-resourced organizations but also those with limited fiscal resources and increased \emph{existential risk} ~\cite{sommer2009managing}? These questions are particularly consequential to future work on ethical and fair practices \cite{crawford2016there}, as these organizations often find applying current best practices in responsible AI development to be too costly. 


To answer these questions, we conducted 17 interviews with practitioners working outside of Big Tech and academia in which we discussed current practices, fairness, and risk mitigation in ML. We used thematic analysis to assess these semi-structured interviews and uncovered six broad themes. 
We explore tensions between privacy and ubiquity, resource management and performance optimization, and access and monopolization. We focus on the subdued impacts of GDPR and privacy legislation, the limited usefulness of model explanations, the trend of deferring responsibility to downstream users and domain experts, and Big Tech's monopolization of access. These tensions reflect underlying and competing concerns of growth and cost, with frequent and complex trade-offs. 

While our findings often overlap with those of past practitioner studies, we find that resource constraints introduce additional challenges to developing and testing fair and robust ML models. Further, even universal challenges of responsible development are exacerbated by an organization's resource constraints---a particularly concerning trend considering ML's growing ubiquity and the rapidly developing support for its democratization.  
Finally, we discuss how the research community can direct future efforts to assist in managing these trade-offs and advocate for giving more research attention and oversight to practitioners in the long-tail of ML development.

\section{Related Work}
Efforts to understand ML practitioners' challenges are commonly assessed through contextual inquiry, surveys, and interviews. While this paper explores the holistic experience of developing, deploying, and monitoring ML systems, past efforts have typically focused on one of explainable AI, responsible AI, or end user requirements. 

\subsubsection{Interpretability}

Much of the literature on ML practice focuses on interpretability. \citet{kaur2020interpreting} conducted a contextual inquiry to explore data scientist use of interpretability tools. Drawing their population from a ``large technology company'' (presumably Microsoft),
they found data scientists over-trusted and misused interpretability tools, and could not generally explain interpretability visualizations. \citet{brennen2020people} conducted interviews with ML practitioners working in a wider variety of contexts. Through their interviews, they observed practitioners in academia and research labs want explanations to provide insight into model mechanics, while other practitioners want explanations that use model outputs ``more effectively and more responsibly''---an understudied use case for explanations. Our study of practitioners at startups, non-tech, and public service organizations reinforced this need. \citet{hong2020human} interviewed ML practitioners who use interpretability methods and found that a core use case for explanations is in building trust both between people and models, and between people within an organization. While our interviews exposed these same desires, few organizations had invested in building interpretability tools, and those that had found only limited success incorporating them.


\subsubsection{Responsible AI}

Responsible AI---particularly focusing on bias and fairness---is another area of ML which has received extensive research attention and contextual inquiry. \citet{holstein2019improving} conducted semi-structured interviews and surveys with ML practitioners at ``10 major companies'' to gain insight into existing practices and challenges in ML fairness. \citet{holstein2019improving} uncovered that---while the ML fairness research community is largely focused on de-biasing models---these stakeholders instead focus on the problems of data collection and data diversity. We found this approach of focusing on data diversity to be similarly common with practitioners from smaller or less visible organizations. 

\citet{rakova2020responsible} introduced a framework for analyzing how company culture and structure impact the viability of responsible AI development. They noted a lack of clarity in how the multitude of proposed frameworks and metrics for responsible AI are translated into practice. When companies introduced such frameworks, practitioners were concerned by the risks of inappropriate or misleading metrics.
Further, they found practitioners at Big Tech companies commonly contend with deficient accountability and decision-making structures that only react to external pressures. We found these experiences and structures to be less applicable for smaller and less visible organizations, where individual contributors generally have more decision-making power, but fewer resources and less developed processes for assessing bias and fairness.

\subsubsection{User Requirements and Expectations}
Another common form of contextual inquiry for ML evaluates how stakeholders will interact with ML systems. To this end, \citet{cai2019hello} interviewed pathologists before, during, and after presenting neural network predictions of prostate cancer, with a focus on their needs for onboarding this technology. They discovered that these stakeholders wanted more insight into the expected model performance: they wanted insight into the model's strengths and weaknesses, as well as the design objective.
Relatedly, \citet{amershi2019guidelines} proposed 18 design guidelines for human-AI interaction, and validated these guidelines with design practitioners recruited from a ``large software company'' (presumably Microsoft). While these design insights are useful for all practitioners, this line of contextual inquiry focuses only on the end human-AI interaction portion of ML development.

\subsubsection{Characterizing Practitioner Needs} 
ML practitioners span diverse fields, industries, and roles. Data scientists are a unique example. This field has been studied separately from ML due to its emphasis on business-adjacent analytics~\cite{trustcorporatedata}, but data scientists are increasingly adopting ML methods. Researchers now study ``Human Centered Data Science,'' which encompasses interest in the shared practices of data scientists and ML practitioners~\cite{HumanCenteredDataScience, muller2021designing, HumanCenteredStudy}. We adopt this broad view of \emph{who} constitutes an ML practitioner, and we do not differentiate data scientists who use ML methods. 

Past research on both ML and data science practitioners categorized various approaches and motivations~\cite{muller2019data, trustcorporatedata, gamut2019} and assessed stakeholders' communication needs~\cite{trustcorporatedata}. These past works highlight two key challenges: the need for continuous refinement, and a desire for clarity in communication and objectives. Data work is inherently ``messy''~\cite{trustcorporatedata}, and practitioners rarely have domain expertise for any given application. Instead, \citet{6674007} found that practitioners learn as they collaborate with new domain stakeholders and grow familiar with the data. The practitioners we interviewed similarly collaborated with domain experts. By and large, interviewees found this process challenging, with communication greatly affected by stakeholders' varied levels of data experience.

\citet{trustcorporatedata} highlighted how teams iteratively negotiated and justified the worth of data science solutions; similarly, \citet{gamut2019} described the \textit{data iteration} processes involved in ML development at Apple. In contrast to \citet{trustcorporatedata}, \citet{gamut2019} focused on functional iteration, updating datasets to improve models and reduce bias, and not iterating on the communication needs of data science and ML teams. Our interviews with practitioners at smaller or less visible organizations exposed similar trends of constant iteration and reinforced the need for communication as an irreplaceable mechanism for addressing the ``messiness'' of data work and for building trust.

\begin{SCtable*}[35][t]
\caption{Overview of interviews, including the type of organization, an overview of the company's main product, the interviewee's title, an interview ID, and an estimate of the companies' overall available resources based on public records and Crunchbase financing reports. One participant's distinctive title has been changed to a functional equivalent to preserve anonymity. In general, \$ companies had less than 15 total employees; \$\$\$\$ had on the order of 100 engineers.}
\label{tab:interviews}
\resizebox{0.78\textwidth}{!}{
\begin{tabular}{lllll}
\toprule
ID & Type  & Company Description  & Interviewee Title  & Resources   \\\midrule
\One & Publicly Listed & Shopping/recommendations & Data Engineer    & \$\$\$\$ \\ 
\Two & Startup &  Shopping/recommendations  & VP of Product           & \$\$  \\
\Three  & Startup & Shopping/recommendations  & VP of Strategy      & \$\$ \\ \midrule
\Four  & Publicly Listed & Pet care (diagnostics)  &  Senior Data Scientist     & \$\$\$\$  \\
\Five & Startup & Healthcare (diagnostics)   & Chief Operating Officer & \$  \\\midrule
\Six   & Startup & Fitness     & Chief Technology Officer                 & \$\$\$  \\ \midrule
\Seven   & Startup & Real estate   & Chief Technology Officer             & \$\$   \\
\Eight    & Small Company & Real estate  & Head Of Analytics               & \$\$ \\ 
\Nine   & Startup & Real estate   & Senior Product Manager               & \$\$ \\ \midrule
\Ten & Startup & ML consulting and tools  & Chief Technology Officer  &  \$\$ \\
\Eleven & Startup & ML consulting and tools & Chief Executive Officer   &  \$  \\
\Twelve & Startup     & Data automation                                       & Board Member/Investor   & \$ \\\midrule
\Thirteen    & Startup & Pet care  & Director of Engineering          &  \$ \\ \midrule
\Fourteen  & Public Sector & Municipality & Asst.~Director of Data Analytics    &  \$  \\\midrule
\Fifteen & Venture Capital & Investment    & Startup/ML Investor              & -   \\ \midrule
\Sixteen & Startup & Language learning  & Chief Technology Officer          &  \$ \\
\Seventeen & Startup & Language learning & Chief Technology Officer           &  \$     \\
\bottomrule
\end{tabular}
}
\end{SCtable*}

\section{Methods}
We used a combination of targeted, convenience, and snowball sampling to invite participants~\cite{etikan2016comparison, emerson2015convenience}: we invited participants by cold-emailing, leveraging our professional networks and the networks of participants. We chose interviewees to represent a variety of contexts, from municipal analytics teams to small startups to publicly traded non-tech corporations. We spoke to a combination of CTOs, directors, investors, engineers, and analysts. All practitioners used or directed the use of ML in their respective work. We categorized interviewees based on available resources; this categorization was inferred through interviews and supported by Public Financial Planning and Budgeting Reports and Crunchbase. We present an overview of these interview contexts in Table \ref{tab:interviews}.


\subsection{Interviewees}
All interviewees did some form of advanced ML development, though these efforts were often pursued for internal tools or as unreleased, experimental products in development. Still, several organizations had released products which used ML methods extensively (\One, \Four, \Five, \Ten, \Eleven, \Thirteen, \Fourteen, \Seventeen, \Sixteen). Every interviewee we spoke with incorporated---at minimum---linear regression methods as a core part of their work, and many used significantly more complex ML techniques. Despite this, we found these organizations often expressed that ``we don’t do ML,” even when ML was  advertised in the company's description, marketing, and website. Several companies rejected our requests for interviews on these grounds. 

Across our interviews, \One{} works at a company most similar to Big Tech: they have extensive resources and advanced engineering practices, and they use ML throughout their business. \Two{} and \Three{} work at an early stage shopping and recommendations company, which is actively migrating to a learned recommendation engine. \Four{} works at a publicly listed, well resourced company that has substantial ML integrations despite not being tech-first.
\Five{} works at an early stage healthcare diagnostics company, and ML is core to their business. \Six{} is a well-funded, late-stage fitness startup with an  experienced ML team, but they use ML in a limited capacity to drive investment. \Seven, \Eight, and \Nine{} work in real estate domains, and all three are relatively early in their ML integrations. Of all our interviews, \Nine{} uses the least ML. \Ten{} and \Eleven{} work at ML consulting and tooling companies; both have strong expertise and advanced practices. \Twelve{} works at a data management company. \Thirteen{} works at an early ML-focused startup with an emphasis on computer vision. \Fourteen{} works in a municipal analytics office; their ML models are typically rapid prototypes. \Fifteen{} supports and invests in ML-focused startups. \Sixteen{} and \Seventeen{} work at early stage language learning companies; both are actively developing core ML features. 

\subsection{Interview Process}
We followed a semi-structured interview protocol. While we adapted our questions for each interviewee, we derived these questions from a set of common themes. 
We sought to survey broader ML and data processes within these organizations---along with the specific challenges these practitioners faced---by eliciting examples, descriptions of existing processes, and anecdotes related to deploying models. We provide the core interview protocols in the supplementary materials, but include samples of questions here:
\begin{itemize}
    \item  What is your process for launching a data-related project?
    \item  How do you evaluate your models or data?
    \item  How do you track projects over time?
    \item  What do updates and changes look like?
    \item  What setbacks have you experienced after launching?
    \item  How do you think about representative data and/or testing?
    \item  How do you think about bias?
    \item  How has GDPR or other data legislation affected you?
\end{itemize}

\noindent Before starting each interview, we obtained permission to transcribe the interview. We intentionally chose not to audio record for interviewee comfort, instead taking notes that included relevant quotes and building transcripts post-interview as described by \citet{rutakumwa2020conducting}. We conducted these interviews by video chat; these lasted an hour on average with the longest lasting two hours and the shortest forty-five minutes. This work is IRB-approved.

\subsection{Analysis}
We systematically coded and analysed transcriptions of these interviews using \emph{thematic analysis}~\cite{braun2006using}. This process involves codifying statements and observations from each interview, grouping statements by initial themes, refining and evaluating themes, and then extracting final themes. The first summary step converted the transcriptions to 945 individual codes. The second step constructed 101 detailed themes. The final step extracted 6 overarching themes. 

\section{Themes}
In reviewing these interviews, we characterize tensions between development and production, privacy and ubiquity, resource management and performance, and access and monopolization. In presenting these themes, we focus on both those sentiments which support prior research analyzing subsets of ML practitioners and those sentiments which are underrepresented in existing literature. 

\subsection{``It's Tough.'' Tensions Between Expectations \& Feasibility}

Human expectations typically manifest as projections of how the model should behave (\Two, \Three, \Five, \Seven, \Eight, \Twelve, \Thirteen, \Fourteen, \Seventeen), and balancing human expectations and feasibility was a recurring theme. Users, leadership, and even engineers often had unrealistic expectations, strong beliefs on the state of ML capabilities, and exceedingly high standards based on prior experiences with other ML products. Practitioners struggled to realize these expectations. 

\subsubsection{Users' Expectations}
Concerns of \emph{user expectations} troubled interviewees. \Sixteen{} and \Seventeen{} quickly learned that users wanted products that performed as well as Google Translate \textit{at a minimum}, but that achieving that level of translational proficiency is infeasible for such early language-learning startups. Failing to meet user expectations would lead to a loss of trust (and ultimately users), but the data, compute, and expertise required to reproduce this proficiency were inaccessible. Both companies had contemplated paying for access to Google Translate's API, but this cost was equally untenable and its use would introduce questions of flexibility and transparency for future development. Instead of using proprietary systems, \Sixteen{} and \Seventeen{} were building their own models and datasets through in-house labeling and user-provided data; both companies ultimately limited the scope of their respective products to reduce risk. For both companies, this decision potentially cost user engagement.

\Eight{} similarly described experiencing ``the big pushback'' when attempting to balance expected user behavior, convenience, and the inherent opaqueness of external, proprietary models and APIs. They considered proprietary models like Amazon's AWS tooling to be ``good products,'' if ``a lot of black box work.'' \Eight{} struggled to balance meeting the users' expectations---for which these proprietary tools are often helpful---with the desire to audit, debug, and maintain control of their software. 
This fundamental tension between meeting expectations---informed by exposure to Big Tech products and models---and managing resource constraints was a concerning and recurring message. These companies found this balancing act to be exceptionally stressful, as they felt user expectations set an untenably high bar given available funding, time, and experience.

\subsubsection{Management's Expectations}
Interviewees found communicating the limitations of ML to be challenging---especially with non-technical management (\Two, \Three, \Eight, \Ten, \Twelve, \Fifteen). For example, \Three{} described a situation in which management identified a seemingly common pattern, but their ML team found themselves unable to extract this ``data story'' given their available resources. Their leadership team identified a common pattern of people first exploring wedding-related content such as dresses, then travel and honeymoon ideas, and finally topics related to child rearing. The team hoped to predict when users began this journey for content curation. In practice, \Three's team was unable to extract this distinctive trajectory from their data, disappointing leadership. \Three{} speculated on the causes underlying this failure: they lacked access to labeled groundtruth to assess predictions and needed additional context \textit{outside} the scope of their data to differentiate signal from noise. They ultimately suggested this normative story of marriage, then honeymoon, then child rearing might be less common in practice.

Some interviewees (\Three, \Ten, \Twelve, \Eight) regarded strong leadership or stakeholder intuition as a warning sign when employing ML tools and techniques. These interviewees found the combination of low data literacy and strong intuition to be most concerning, and remarkably common in both startups and non-tech companies. As {\Twelve } explained, ``CEOs and executives don’t really understand what it takes [to develop and deploy ML],'' particularly outside tech-first organizations. One interviewee (\Ten) declared that in cases of low stakeholder technical and data literacy, they opted not to pursue  contracts to avoid wasting time and resources. 



\subsubsection {Predicting Performance and Cost}
The ML community is increasingly aware of \emph{model under-specification}, one symptom of which is that training the same model multiple times on the same data results in different defects~\cite{d2020underspecification}. Practitioners---particularly in under-resourced environments---are patently aware of this challenge (\Six, \Thirteen). At Big Tech companies, a common mitigation strategy is to train a model multiple times on the same data, assess the resulting 
models, and deploy the best of the bunch. In our interviews, only the most technology-first company (\One) adopted this approach. In line with other less-resourced organizations, {\Thirteen } discussed how finances expressly prevented them from doing so. 

{\Thirteen } shared that they did not have the resources to train multiple models for the same set of data. As they scaled their business, they collected more data---in their case, data which mostly consists of videos of dogs. With each new data collection iteration, they retrained their model using all available training data, resulting in unpredictable performance drops. They had been debating decomposing their model into simpler models corresponding to specific characteristics, such as dog breed. In this manner, they could scale their product by introducing new models without the risk of compromising past performance on other breeds. Of course, the downside is substantial: instead of testing a single model, they would need to test many. Further, they believed the larger model would be more robust as it should better generalize to underrepresented data---such as mixed breeds, or atypical characteristics---whereas breed-specific models would be substantially worse. 

In response to these challenges in developing models and predicting ML development costs, many interviewees (\Six, \Seven, \Nine, \Fifteen, \Twelve, \Fourteen) simply wondered whether deploying ML models was worth the trouble at all. One interviewee (\Seven) had concluded it was not. In reference to to their work in real estate they stated, ``I don’t have any worries about automating the quality checks,'' but ``we’ll never have the density of data to really automate some of these things... the ROI [Return On Investment] on automation might be low.'' After extensive preliminary development, \Seven{} came to believe that human intuition and domain knowledge was irreplaceable---models make predictions based on known prior behavior, but domains like urban property appraisals are inconsistent, constantly fluctuating based on ever-changing human priorities.

\subsubsection{Discussion: Expectations and Feasibility}
Practitioners struggled to meet human expectations---whether from users, from leadership and management teams, or even from themselves. User expectations are dynamic, but are largely informed by the practices of Big Tech companies (\Sixteen, \Seventeen). When extremely well-resourced organizations release an ML product into the world, it raises the bar of expected quality and effectively \textit{monopolizes access}. To participate, organizations must opt in to use of released, opaque models and APIs or invest beyond their means in collecting data and modeling. 


The lore of ML has resulted in non-technical management believing that pattern recognition should be straightforward, but practitioners (\Ten, \Three) often find themselves unable to meet these unrealistic expectations. When we researchers talk about AI and ML, and especially when we engage popular press in these conversations, we establish expectations about the capabilities of ML. As we release new models, tools, and techniques, we set user expectations for feasibility and quality standards. Small and less-resourced organizations struggle to meet these expectations. With this monopolization of access, many questions arise: if organizations use proprietary models downstream, how do they introduce transparency or maintain autonomy? Who is responsible? And what are the implications of these participation monopolies? 

When developing models internally, interviewees cited several potentially viable approaches to meeting these expectations, such as replacing large, complex models with multiple, specialized models targeting specific tasks (\Thirteen). But this is a relatively new topic, and practitioners would benefit from guidance on \textit{when}, \textit{how}, and \textit{whether} to transition from a single model to multiple  specialized models. Understanding the costs and benefits for such transitions is crucial as well---what are the monetary and environmental expenses involved in training and evaluation for each choice~\cite{strubell2019energy}? 

Further, while we often think of ML as a useful and perhaps inevitable tool, these organizations question that narrative (\Six, \Seven, \Nine, \Fifteen, \Twelve, \Fourteen). The decision to implement ML typically falls to management and team intuition, and is sometimes merely employed as a mechanism to drive investment (\Six, \Twelve). This community of long-tail, less visible ML practitioners would benefit from standardized recommendations for assessing when ML is most appropriate.

%
%
\subsection{``A Hotbed of Bias.'' Efforts to Assess, Prevent, \& Mitigate Bias}
Nearly every organization we interviewed expressed substantive concern over inadvertently deploying biased models (\One, \Three, \Four, \Five, \Seven, \Ten, \Eleven, \Thirteen, \Fourteen, \Sixteen, \Seventeen). \Eleven{} stated this to be the ``biggest business concern'' for companies incorporating ML, and other interviewees shared similar sentiments (\Sixteen, \Seventeen). Yet strategies for uncovering and mitigating bias and its impacts were consistently underdeveloped, pointing to the potency of this problem.

\subsubsection{Bias Mitigation Through Diversity or Personalization}
A remarkably common mitigation attempt to alleviate bias was through the acquisition of sufficiently diverse data for model training and evaluation (\Sixteen, \Five, \Eight, \Ten, \Three, \Thirteen, and \Fourteen). This strategy contrasts with academic approaches which aspire to debias models assuming fixed data, but mimics the broader practices of Big Tech~\cite{holstein2019improving}.
Mechanisms for acquiring this diverse data include attempting to develop a sufficiently diverse userbase (\Sixteen), ingesting data from varied sources (\Eight, \Ten), augmenting available data (\Eleven), collecting diverse data in-house by assessing \emph{axes of diversity} (\Thirteen), and incorporating participatory design principles directly into data planning and collection mechanisms to ensure representational data (\Fourteen). Another commonly proposed bias mitigation strategy considered model personalization (\Seventeen, \Sixteen, \Two, \Thirteen). By giving users increased control over their models, these companies argued that their users could tailor outcomes to their personal needs---circumventing the problems of biased models, or so the logic goes.

Still, data-focused mitigation strategies suffered from drawbacks: practitioners developed ideas of how models would behave on a given dataset, but after each modification or retraining found the resulting performance to be unpredictable. As such, \Thirteen{} adopted a slow and cautious update protocol: they avoided introducing new, diverse data until user requirements deemed it strictly necessary. By doing so, they minimized how frequently they updated---and potentially broke---their models. \Thirteen{} pursued this strategy as their product was in Beta, and they found it offered greater consistency. For a deployed model, though, this slow and cautious approach to adding new and diverse data can exacerbate or prolong model biases. Despite the increased risks, this slow and cautious strategy to data changes is common in production: teams frequently use model accuracy to appease investors (\Five, \Fifteen), which may lead to them opting to deprioritize data diversity to reach investor expectations. 


In spite of their varied bias mitigation strategies, interviewees nonetheless remained concerned about releasing biased models. They lamented  they had no systematic way of acquiring diverse data (\Thirteen), of assessing the axes of diversity (\Ten, \Thirteen), or of assessing and evaluating the performance of personalized models (\Seventeen, \Sixteen, \Two, \Thirteen). For example, \Seventeen{} explained they were considering investing in federated ML for personalization---but assessing the quality of each personalized model is hard~\cite{vanhaesebrouck2017decentralized}, as is the engineering challenge behind such a strategy. In two separate vision applications, \Thirteen{} and \Six{} further lamented that the scope of diversity needs was \textit{far greater} than they initially anticipated: they must consider not only diversity of image subjects, but also the diversity and quality of seemingly unimportant image backgrounds. 

\subsubsection{Assessing Blind Spots}
A related concern is creating models which suffer from \emph{blind spots}: undetected failures due to missing or biased data and inadequate test coverage. Many interviewees identified human subjectivity in data collection and labeling consistency as the root cause of these failures (\Eight, \Fifteen, \Seven, \Three, \Eleven), contrasting in part with the research community's broader concerns that models do not learn the causal relationships within an available, finite dataset~\cite{d2020underspecification}. Human subjectivity affects the full development pipeline from deciding what data to collect to data labeling to model assessment and evaluation. Some interviewees expressed optimism about efforts to use internal quality metrics as mitigation mechanisms to find and ultimately remove these blind spots (\Fifteen, \Seven, \Eleven, \Three). \Eleven{} was similarly optimistic that by holistically considering their data and performance, they would be able to isolate data factors related to computational \emph{underperformance}. 

Still, interviewees were optimistic that ML could be deployed responsibly: they believe model biases to be more scrutable than human decisions (\Seven, \Fourteen, \Five). One interviewee (\Five) explained that because they drew training and test distributions from the same imaging devices, they believed they strongly upheld the assumption that training and test data are independent and identically distributed. Further, \Five{} had collaborations with insurance companies; they argued this resulted in increasingly representative data. As a consequence, they asserted that they were insulated from blind spots. Interviewees expressed desires for better assessments of blind spots and model fairness (\Ten, \Sixteen, \Thirteen, \Fourteen).

\subsubsection{Deferred Responsibility}
Several interviewees expressed that biased models were a possibility, but that these models could still be deployed safely by leveraging a final barrier of human judgment (\Eight, \Five, \Thirteen, \Eleven, \Fourteen): when faced with an incorrect assessment, a human would be capable of overriding the model's decision (\Five, \Eight, \Eleven) or disusing the model (\Fourteen). We find this trend of deferred responsibility to be common among these practitioners and troubling. Past research has demonstrated that even when a human is capable of outperforming a model when acting independently, they tend to defer to model predictions when available~\cite{dzindolet2003role,suresh2020misplaced}. One interviewee (\Four) expressed concern about this possibility of unintentionally undermining human expertise. To safely deploy their models, \Four{} was actively collaborating with a design team to emphasize decision uncertainty. However, several other practitioners adopted a notion that building tooling to support downstream users and organizations was unnecessary, believing that users were sufficiently capable without this hand-holding (\Eleven, \Five)---a risky position. This highlights the murky notion of responsible parties: are developers responsible for the consequences of introducing ML, or are end users and domain experts who use these models?

\subsubsection{Discussion: Assess, Prevent, \& Mitigate Bias}
Risk and harm assessments were recurring themes. When interviewees deferred responsibility to others, they assumed their products could cause minimal harm. When practitioners considered risks to be sufficiently low, they felt little responsibility to consider the potential for harm. These risks are not distributed equally: prescribing home valuations (\Seven) is demonstrably more risky to human wellbeing than ranking ``cool'' shirts (\Three). Still, this is a slippery slope of complacency. We assert assumptions of minimal risk are inherently biased and not appropriately calibrated against feedback from users and stakeholders. Even a shirt recommendation can cause harm and perpetuate inequality---e.g., if the model consistently demotes Minority-owned brands. While many interviewees expressed concern for the broader implications of their work (\One, \Two, \Three, \Ten, \Eleven, \Thirteen, \Fourteen, \Seventeen), several did not (\Nine, \Six, \Eight, \Twelve, \Five). This apathy was exacerbated when practitioners relied on humans to arbitrate decisions. \Five{} explained that blind spots were ``low risk,'' with bad outcomes resulting in a ``\$300 procedure instead of a \$50-60 procedure,'' but the person receiving this treatment might not agree. 

Trade-off considerations between robustness and accuracy were a common theme.
Instead of the intuitive or ad-hoc methods that our interviewed organizations currently use to assess risk and mitigate bias, we suggest developing methods to recognize risks and harm through atypical perspectives---such as by characterizing expected users and then assessing any embedded biases within these expectations for rectification. Some companies bring in consultancies to do this, but this is untenable with resource constraints. We assert that the research community can help by designing tools which actively encourage critical thinking, monitoring, and assessment for data planning and modeling tasks~\cite{mehrabi2019survey,mitchell2019model}. Such efforts can better support these less-resourced organizations.



\subsection{``You can poke and prod black box models, right?'' Black Boxes \& Overconfidence}

Concerns of bias are often associated with ML techniques operating as ``black boxes''~\cite{lipton2018mythos}. Practitioners had a broad swathe of opinions and strategies for managing, using, evaluating, and avoiding black box models. Two comments were particularly surprising. First, practitioners explained that ML models are not the only black boxes they engage with---so this is familiar territory. Second, practitioners were more optimistic about engaging simple prodding-and-probing methods to understand black boxes than previously reported~\cite{hong2020human}. 

\subsubsection{Black Boxes, Explanations, \& Transparency}
Prior work has extensively engaged with concerns of learned black box models. These works point to widespread desires for explanations \cite{hong2020human,kaur2020interpreting,brennen2020people}. We instead found practitioner opinions on ML explanations vary widely. We found \emph{ambivalence} to be a common sentiment toward using black box models, particularly when the perceived potential for harm was low (\Two, \Three, \Five, \One, \Thirteen, \Six). For example, \Two{} stated they ``don’t mind the \emph{unknown} of black box,'' though they ``do mind if it causes harm.'' \Five{} stated explanations were not necessary as their product only \textit{supports} a human in diagnosis---so, again, the black box is low risk. Similarly, \One{} felt little concern using black box models, with the caveat that ``From an ethical perspective, I don’t think one should write a model if they have no idea how to tell when something goes wrong.”  

Many interviewees (\Two, \One, \Nine, \Four, \Three) pointed to the usefulness of \textit{transparency through example}, where examining sets of inputs and outputs provided sufficient insight into the model behaviors \cite{booth2021}. \Three{} wanted awareness for how a model reached its goal through increased cognisance of the impacts their objective functions. They detailed how, after adapting their recommendation model to optimize for click-through rate, their model began recommending fuzzy pictures and ``weird'' content. Despite fulfilling their objective, the content was not useful to their users. To address this, they introduced more stringent prodding-and-probing processes. While \citet{hong2020human} suggested this style of proding-and-probing experimentation did not meet practitioners' needs for transparency, we instead discovered that interviewees found this process of experimentation and rigorous documentation and versioning mitigated their concerns about using black box models.

In addition, \Four{} and \Ten{} desired classic ML explanations. \Four{} wanted to provide explanations to help users assess predictions, believing this especially necessary for calibrating trust. For internal interpretability use, \Four{} had implemented LIME~\cite{ribeiro2016should} and feature importance methods, but found these explanation methods unhelpful for their end users. \Ten{} also desired explanations, stating ``nobody wants a black box model.'' As an intermediate solution, \Ten---a consulting company---relied on Shapley Values \cite{shapley1953value} to increase customer trust, though they noted it was not an ideal solution. They cited two challenges: first, for each data type, they needed to design a custom interface for providing these explanations. Second, they found their users did not know how to interpret these values.

Finally, \Eight{} presented a unique perspective on black boxes, stating that ``black box'' could equally describe teams, processes, decisions, proprietary models, and obfuscated APIs. \Eight{} found these non-ML black boxes to be equal sources of frustration. They opted to stop using proprietary models as they found these hard to inspect when problems arose. Similarly, undocumented API changes often introduced data errors. Even people could act as ``black boxes'' when not effectively communicating, and described letting go of an engineering team that wouldn't explain their work. Ultimately, \Eight{} saw black boxes---whether models, processes, or people---as a tool to avoid. 

\subsubsection{Efforts to Mitigate Overconfidence}
Overconfidence is a common ML problem wherein a model's average confidence greatly exceeds its average accuracy~\cite{guo2017calibration}. This is related, in part, to the question of deferred responsibility: companies often sought to catch errors by having humans assess low confidence predictions (\Thirteen, \Five, \Four). In our interviews, discussions of overconfidence did not focus on strategies for calibrating the confidence of models, as is trendy in research~\cite{guo2017calibration}. Instead, interviewees typically described user-facing mitigation strategies such as onboarding or uncertainty communication. 
For example, \Five{} relied on onboarding users for their predictive healthcare models and explained how healthcare practitioners should interpret their confidence metrics. They stressed these confidence metrics are \emph{not} probabilities---100\%\ confidence does not indicate certainty, nor did 50\%\ randomness. These proprietary scores were designed with user feedback and practitioner experience. The team also developed sensitivity thresholds for categories of healthcare practitioners, noting these practitioners tended to have higher or lower tolerances for false positives. During their onboarding process, they explained these sensitivity thresholds and when to anticipate false positives or negatives. 

In contrast to \Five's approach of relying on onboarding to mitigate overconfidence, \Four{} sought methods to continuously calibrate user trust and understanding. To do so, \Four{} collaborated with designers to explore better presentation of uncertainty and curated model explanations. In working with these designers, they expressed a desire to present information in a ``non-definitive way.'' We argue that the former approach of using onboarding and education to mitigate overconfidence is unlikely to be effective: prior research in the explainable AI and human factors communities has shown that managing human responses to automated decisions is a major challenge \cite{dzindolet2003role,suresh2020misplaced}. Presenting information in a non-definitive way has more promise, but uncertainty communication remains a well studied yet unsolved problem \cite{deitrick2006influence, padilla2020uncertainty}.

\subsubsection{Discussion: Black Boxes \& Overconfidence}
Some interviewees described black boxes as part of life---unavoidable, perhaps frustrating, but certainly normal. Despite the ambivalence some interviewees (\Two, \Three, \Five, \Fifteen, \Nine, \Thirteen, \Six, \One) displayed when discussing automated explanations for black box systems, others (\Four, \Eight, \Ten, \Eleven) were deeply worried about their implications. While these companies were willing to use these black box models as a supporting measure---particularly for things humans are bad at, like repetitive tasks (\Thirteen)---they were not willing to substitute ML in place of human intuition. \Seven{} explained, ``I don’t have any worries about automating the quality checks.” However, they also said this of their higher risk property assessment use case: ``the value of something is not objective. If you had a perfect algorithm for what something is worth, but then someone buys it for three times as much, then that’s what it’s worth.'' According to \Seven, no amount of transparency could make these techniques sufficiently safe in this context---they would always be conducted by a human.

In the explainable AI community, there is a latent assertion that poking and prodding a model can never be sufficient for understanding a model; instead, we need a mathematically-meaningful yet to-date-indeterminate mechanism to provide actionable insights into the decision-making mechanics of ML models. Past research with practitioners from Big Tech has confirmed a desire for this form of meaningful explanation method \cite{hong2020human}. However, many of the practitioners we interviewed (\One, \Two, \Nine, \Four) expressed more optimism about the idea of transparency by example, where examining sets of inputs and outputs allows developers or users to build accurate mental models of the ML model. In contrast with past research, these practitioners believed that adopting this testing procedure could suffice for model understanding.

Still, some interviewees did desire explanations (\Four, \Ten); their desires reflected insights from \citet{hong2020human} and \citet{gosiewska2019ibreakdown}. These practitioners found explanation methods to be unstable and did not generalize well to their contexts. Other interviewees (\Two, \Four, \Ten) desired explanations for non-technical stakeholders and users when uncertainty and overconfidence were concerns. But these explanations are costly to set up---requiring carefully crafted user interfaces and broader tooling---and are still very limited. \Four{} explained that the methods they had tried---LIME and feature importances---were insufficient at communicating uncertainty, reflecting recent work by \citet{bhatt2020uncertainty}. This similarly aligns with prior research on uncertainty communication showing shown numerical representations are not effective at this, nor are confidence intervals \cite{1231171,errorbars}. We suggest future work in explanations emphasize uncertainty communication and participatory design processes \cite{muller1993participatory}, as there is a desire for this work and rich prior literature in Data Visualization and HCI communities \cite{padilla2020uncertainty}.


Onboarding for ML products and tools is an underexplored area. \citet{cai2019hello} detailed how clinicians benefit from onboarding covering the basic properties of a learned model, such as known strengths and weaknesses, its development point-of-view, and design objectives. \Five's approach overlaps with these recommendations, but could benefit from increased transparency and focus on positioning the model's capabilities with respect to the downstream domain expert. Like \citet{cai2019hello} suggest, we believe the onboarding process for human-AI teaming should be continual and evolving, and should re-address use and understanding of tools over time. Onboarding should be designed based on studies of how users interact with the model, and should progress as users' mental models develop through use and exposure. 
In particular, any time an AI tool is updated---especially considering that updates may break human expectations of model performance without warning~\cite{bansal2019beyond}---onboarding should be readdressed.

%
%
\subsection{``Data Literacy Is Not a Silver Bullet.” \\ On Communication \& Collaboration} 

As documented in prior work, ML practitioners do not work in isolation~\cite{trustcorporatedata}. Instead, their projects typically require collaborations with domain experts and other cross-functional collaborations. Every interviewee raised effective communication as critical to their work, and \Three{} described this simply as ``really tough.'' Despite the impactfulness of effective communication, best practices were \emph{never codified}, and instead comprised of unwritten institutional knowledge and ``water cooler talk'' (\One).

\subsubsection{Effective Communication}
Interviewees cited communication as the principle mechanism to reduce and mitigate the risks of using ML (\Four, \Five, \Eight, \Seven, \Fifteen, \Ten, \Eleven, \Fourteen, \One). Unsurprisingly, when communication was lacking, practitioners found they wasted time and money on disagreements (\Eight) and inappropriately targeted work (\Eleven, \Fourteen, \One). Several interviewees discussed data literacy as a bottleneck to effective communication (\Eight, \Ten, \Fourteen). \Eight{} discussed how mismatches between human expectations of business narratives and the analytics led to data insights being ignored: ``A couple people in the company have a narrative in their head already, and if the analytics don’t support that narrative, they don’t want anything to do with those analytics.'' When stakeholders do not understand the metrics, they also do not \textit{trust} the output---resulting in wasted effort (\Eight, \Fourteen). In response, \Eight{} routinely engages in negotiations with business teams, often requiring external mediation to resolve disputes. Similarly, \Fourteen{} lamented the ``politics around the numbers'' and how these so-called politics undermine their work. However, \Fourteen{} also cautioned that a domain expert who was \emph{too} technical was equally challenging to establish a productive working relationship with: instead of expressing their needs, these technical stakeholders spend time anticipating potential analytics pitfalls.

\subsubsection{Institutional Knowledge}
Over the life cycle of a project, ML practitioners become more familiar with their data, models, and domain. This evolving awareness affects data collection and model iteration but is typically shared person-to-person, not systematically. For example, \Ten{} tracked large data and model changes through ``informal discussions,'' while \Eight{} ran scripts on an ad-hoc basis to check for data quality errors they noticed over time. Similarly, \Seventeen{} described learning to intuitively assess the importance of errors based on past user feedback. These insights were rarely documented, but were often embedded in code and shared through conversations. We refer to this as \emph{institutional knowledge}: information that exists across a team but is not recorded. Existing literature charmingly describes institutional knowledge as ``the stuff around the edges'' \cite{mccurdy2018framework}, considering the context, history, background, and knowledge entwined in artifacts like data and models. Yet, this raises questions of project permanence and responsible development: undocumented information will likely be forgotten. Only \Fourteen{} described efforts to explicitly document this acquired knowledge, citing metadata as their primary tool for explicating changes to datasets. We speculate that this effort is a consequence of their team's positioning within a public municipality office. 


\subsubsection{Discussion: Communication and Collaboration}
Developing stakeholder alignment is not a new challenge, but ML and data contexts introduce new gaps in vocabularies and experiences.
Tools that assist non-technical stakeholders might bridge these gaps through by supporting data literacy and learning \cite{hopkins2020visualint}. Practitioners currently struggle to bridge these gaps, but doing so is critically important, particularly when practitioners are not domain experts and so need stakeholder insight for new features \cite{6674007}. 

Relying on institutional knowledge is also not unique to ML development. Nonetheless, we see this as an area which lacks sufficient investment both in research and in practice, and which could be improved through the introduction of comprehensive standards~\cite{bender2018data,mitchell2019model,holland2018dataset}. An added benefit of more consistent artifacts is the increased potential for internal auditing and quality checks during development~\cite{raji2020closing}. Even beyond ML development, practitioners need careful consideration of the value of ML development to answer \textit{is it necessary?} Future work should produce tools for comprehensive social-systems analyses  encouraging stakeholders to examine the
possible affects of models on \textit{all} parties. 


%
%
\subsection{``Experiment, Iterate, See We’re Getting Closer.” A Model Is Never Finished}

The life-cycle of a ML model requires continuous refinement. This is part and parcel of both an organization's growth and practitioners' increasing awareness of model limitations. Practitioners' response to challenges in planning, iterating, and evaluating are varied and often reactive. These responses are best characterized by what one interviewee called a ``lack of best practices in training'' (\Eleven). Practitioners sought to improve their work despite resource constraints, operating under yet another \textit{catch-22}: investing in reflection and refinement while staying within budget. 

\subsubsection{Data Quality: Planning, Ingesting, \& Cleaning}

In line with existing literature \cite{gamut2019}, we found that ML development is significantly hampered by challenges in data collection, management, and use. Nearly every company struggled to standardize data entry, collect enough data, and collect the \textit{right} data to both mitigate bias and encourage robustness. \One{} was the only exception, as they worked primarily on ML tooling and not immediate business applications. 
A major challenge in building ML models was predicting data requirements for good performance: interviewees complained their initial estimates were substantially incorrect, and explained that they actively seek to collect more and additional sources of data (\Four, \Twelve, \Eight, \Three, \Thirteen, \Six). Understanding data coverage needs by assessing real world variability and translating this to data requirements remains an open question in the ML community.

Because of this open question, practitioners often developed a reliance on a subset of \emph{trusted data} (\Twelve, \Seventeen, \Nine, \Three, \Fourteen, \Six). \Three{} explained that they scoped their recommendations based on the quality and consistency of the available data. They tried adding recommendations for brands with worse data cleanliness, but ``it affects the models, because every bit of data you have is fragmented across hard-to-reconcile records.'' \textit{They were not willing to take on the risks that messier data introduced to their recommendations.} Instead, they relied on data from a small set of familiar brands, explaining that while they considered promoting diverse brands, investing in data cleaning was ultimately too costly given the company's precarious position in entering the market. Lastly, companies found that even in labels provided by domain experts, labeling inconsistencies and disagreements were common problems. In response, several companies developed complex routines for building consensus (\Four, \Five, \Thirteen, \Eight). These findings support work by \citet{gamut2019}, emphasizing that challenges in data management are universal but are nonetheless often deprioritized when facing resource constraints. 

\subsubsection{Many Methods of Evaluation}
In all of our interviews, we discussed the immense challenges of model evaluation extensively. No two companies had the same process, but every process involved multiple evaluation mechanisms. Above all else, extensive manual evaluation using hand-selected test cases was key to these strategies (\Seven, \Sixteen, \Four, \Eight, \Five, \Seventeen, \Three, \Thirteen, \Eleven \One). This first evaluation step was described as ``weak but useful'' by \Three. Beyond manual evaluation, many companies implemented supplemental A/B testing (\Three, \One) or beta tests with in-the-wild users (\Sixteen, \Four, \Seventeen, \Thirteen, \Fourteen, \One, \Six). Several interviewees discussed the tradeoff between using extensive in-house evaluation and relying moreso on user feedback (\Thirteen, \Eleven, \Seventeen, \Sixteen, \Two, \Fifteen, \Seven): user feedback is significantly cheaper and contains additional signal on whether errors are impactful to users, but mistakes cost user trust and engagement.  

Interestingly, \Four{} compared beta testing models to their clinical trials with veterinarians. They considered this process necessary partly due to the black box nature of ML models, but also because they found it to be useful for assessing the broader impacts of deployment. They emphasized a desire to bring the rigor of clinical trials to their model evaluation---for example, through randomized control trials assessing the introduction of models to veterinary businesses. \Ten{} also proposed adopting the scientific processes to assess ML by formulating and testing hypotheses. Both sentiments follow proposals to adapt the scientific study of behavior to ``intelligent'' computational models~\cite{rahwan2019machine}.
Lastly, none of the interviewees indicated that they had an effective methodology for evaluating fairness---though several expressed this as a desire (\Fourteen, \Ten, \Thirteen), and \Ten{} was exploring Model Cards~\cite{mitchell2019model} as a step in this direction.

\subsubsection{Model and Data Versioning} Many interviewees desired better model and data versioning (\Two, \Fifteen, \Ten, \Five, \Eight, \Seven, \Three, \Thirteen). Some companies pointed to recent distribution shifts caused by COVID-19 as highlighting its importance (\Seven, \Eight, \Four). Still, versioning remained elusive, and \Seventeen{} explained their company was ``too early'' to invest in it.
Four companies (\One, \Four, \Thirteen, \Eight) did extensive model versioning. \One{} included complete versioning of all the ``data that went in, and the code as well'' as a component of their evaluation pipeline. \Thirteen{} and \Eight{} version ``everything,'' and \Thirteen{} explained they're only able to afford this process because of a Google Cloud credit award. Were it not for these credits, they would not have the storage capacity needed to version their models and datasets.



\textit{Metadata} is critical for evaluating data, directing modeling iteration, documenting changes, and retroactively incorporating data and model versioning (\Fourteen, \One, \Twelve, \Two, \Three). \Fourteen{} explained they relied on metadata to identify ``what worked within a dataset'' as metadata can reflect ``things that are taken care of as the project is going on.'' Similarly, \One{} relied on metadata to inform their work, even training overnight based on ``metadata considerations.'' When creating groundtruth, \Twelve{} explained their central question is, ``what metadata do I need to make sure this product matches the end description?'' Metadata serves as documentation for institutional knowledge, 
yet remains an underutilized resource. 



\subsubsection{Discussion: A Model is Never Finished}
Each interviewee adopted different processes for evaluating their models. This is not surprising: mechanisms for effectively testing ML models remain rudimentary, and recommendations inconsistent. We assert the research community should produce consistent testing recommendations, with an increased focus on test cases \cite{ribeiro2020beyond, booth2021} and model fairness assessments~\cite{mitchell2019model}.
Prior work described comparing multiple models as \textit{crucial} during evaluation \cite{hong2020human}, yet we found interviewees rarely implemented this process---while it is common to train models on different subsets of data, hyperparameters, or with different seeds \textit{in parallel},  resource constraints can make training multiple models for comparison impossible (\Seventeen, \Thirteen, \Eight, \Sixteen). \Four's proposal of using randomized controlled trials and other processes adopted from clinical trials to assess models is compelling. We believe the research community should further recommend best practices for adapting these evaluation mechanisms for ML contexts~\cite{wiens2019no}.




ML experience levels affected development practices. In cases where teams lacked ML experience, poor modeling decisions were followed by periods of indecision (\Twelve). In some cases, these teams might end ML development entirely (\Twelve). More mature teams (\Six, \Ten, \Fourteen, \One) emphasized the fleeting lifespan of models and encouraged team members to prioritize frequent retraining, minimally complex models, and a willingness to ``throw it out'' (\Six). In contrast, less experienced organizations might not \textit{afford} retraining and replacing models, or lacked the experience to build modularity into models. As such, \Three{} characterized ML development as notably slow and defensive compared to other engineering tasks. We were reminded of parallels in software development best practices: guidelines made code cleaner and easier to debug or replace \cite{parnas1979designing}. We believe future work adapting these practices for ML development and maintenance would be beneficial to practitioners.


%
%
\subsection{``GDPR Doesn’t Affect Us.” Assessing Tensions Between Privacy \& Growth}

Machine learning necessitates the collection and management of large data collections. As a consequence, recent legislation such as GDPR and the CCPA have broad implications for the discipline. We asked interviewees about their relationships to these and other privacy legislation works to assess how practitioners are responding.

\subsubsection{Government Regulation \& Privacy Policy Impacts}
The academic community continues to debate if GDPR encompasses the \emph{right to an explanation} for an automated decision, but collectively agrees that GDPR encodes at least the \emph{right to be informed}~\cite{goodman2017european}.
While the former interpretation has stronger implications for ML practitioners, both should have some affect. GDPR is complemented by emergent legislation seeking to protect user privacy---with many implications for data collection and handling~\cite{stallings2020handling}. We asked ML practitioners how this legislation affected their practices. Though \Ten{} noted their company's legal team guided them on GDPR and other regulations, \emph{no} interviewee indicated that they were directly concerned with the requirement to provide an explanation, despite using black box models extensively. By and large, interviewees expressed that GDPR and other  legislation had not impacted their work in any substantive capacity (\Two, \Nine, \Sixteen, \Ten, \Four, \Eight, \Three, \Eleven). 

A few interviewees explained that, to comply with GDPR, they leveraged their nature as platforms to avoid collection of personally identifiable data (\Four, \Ten, \Eight). Others lamented minor inconveniences relating to GDPR, such as increased latency from using remote servers based in Europe (\Ten) or vague concerns over future implications for cloud use (\Eleven). 
One interviewee (\One) described how their company responded to GDPR by extensively changing their data handling practices, devoting over six months of engineering time to ensuring compliance. We should note that this company was both well-resourced and publicly listed. While they bemoaned that deletion requests do come in, and are ``a pain in the ass,'' the interviewee explained that the process of adopting GDPR compliance in data handling had actually been immensely beneficial to the company. GDPR forced their company to develop better awareness of and practices for handling data, and this re-evaluation increased their overall data competencies. Despite implementing these extensive changes to data handling, \One{} nonetheless remained unconcerned with any notion of a ``right to an explanation.''

\subsubsection{Privacy Legislation is Insufficient} A common sentiment was that privacy legislation continues to be insufficient to protect users. Interviewees often felt it necessary to implement their own policies and tooling beyond any requirements (\Seventeen, \Sixteen, \Thirteen, \Three, \Six). In the absence of stronger privacy legislation, companies aspired to act ``without malcontent whenever possible'' (\Three). Companies continue to internally assess their responsibilities to users' privacy, but find themselves attempting to balance these responsibilities with other desires. Many companies discussed managing the tension between user privacy and their desires to become ubiquitous, and to collect ever more extensive datasets (\Sixteen, \Seventeen, \Thirteen, \Two, \Three, \Six).  



\subsubsection{ML to Satisfy Regulators}
While we asked interviewees about how legislation changed their ML development and data practices, one interviewee explained that they instead used ML to respond to regulatory requirements (\Seven). In their real estate assessment business, regulations require reporting on properties' conditions and features. While their ML models had insufficient accuracy to meet consumer expectations, they found this accuracy rate to be acceptable for ensuring broad-sweeping regulatory compliance. 

\subsubsection{Discussion: Tensions Between Privacy and Growth}
The relationship between privacy legislation and ML development is curious. From the researcher's perspective, the world is abuzz with chatter about the implications of GDPR for explanations of automated decisions. Yet, \emph{every} practitioner we interviewed was unaware of these discussions, let alone the need to revise ML development practices in response. GDPR and other privacy legislation had started to affect data practices, but ambiguity abounds:  while one company implemented extensive changes to their data management systems (\One), none of our interviewees considered the implications of these deletion requests for ML models. Should this deleted data also be deleted from any model training, test, and validation sets? Should the \emph{model itself} be deleted in response to the request~\cite{ginart2019making,villaronga2018humans}? These questions go unanswered in the research community, and unnoticed in these practitioners' realms. Organizations (\Sixteen, \Seventeen, \Thirteen, \Two, \Three, \Six) continue to self-moderate ideas of ``acting without malcontent'' (\Two)---analogous to Google's antiquated motto of ``Don't be evil.'' These organizations experience tensions between their desires to sustain and grow their businesses, and to protect user interests.


\section{Conclusion} 

When discussing ML practice with smaller and less visible organizations, we find these practitioners have many commonalities with their Big Tech counterparts: desires for explanation, lack of standardization, and unending difficulties communicating with stakeholders. We also uncover several new, divergent findings: for example, in contrast with past studies, the practitioners we interviewed expressed optimism for transparency through example, noted access differentials for bias mitigation, and experienced subdued implications from privacy legislation on ML development. 

Most critically, resource constraints affect the development of responsible ML and amplify existing concerns about the challenges of responsible and fair ML development for these interviewed organizations. These constraints continuously affect the work that companies and organizations invest in; for example, while several companies wanted to invest in explanations for ML models, they found the costs of developing these techniques to be too high, especially given the ``researchy'' nature of these tools. Intuitively, the resource constraints of startups and small companies encourage increased caution in decision-making, but this requisite careful planning is untenable without sufficient experience and domain knowledge---both of which are difficult to acquire. Instead, organizations found predicting ethical and financial costs to be difficult, causing them to reconsider incorporating these methods. New (ML) product explorations were accompanied by exploding budgetary requirements. \Six{} suggested that, as with other forms of engineering work, predicting costs for ML requires experience and exceptionally large buffers of both time and money due to increased risk or complexity. But many of these companies lack the necessary expertise to assess cost and found the ``cost of going external [to be] too high'' (\Eight). These buffers of time and resources represent untenable costs for organizations actively seeking investment. 

Finally, assessing the social implications and necessity of modeling is important but risky---in Big Tech teams, misunderstandings and oversights do not collapse the business. Big Tech has large teams with expertise in data and ML, as well as extensive investment in in-house tooling. In contrast, many interviewees lamented the challenge of hiring ML talent (\Four, \Eight, \Twelve, \Fifteen, \Two, \Seventeen, \Sixteen), and the lack of accessible and comprehensive tools to assist in much-needed bigger picture analyses. One interviewee suggested less-resourced organizations' products are especially prone to ``exhibiting this bias'' \textit{because} of limited resources and expertise (\Eleven); all the more reason to center these practitioners as we consider the challenge of responsible ML development.


The potential benefits of ML technology must be spread beyond the agenda of Big Tech and into all corners of society, yet the vanguard of small organizations implementing ML struggle to realize the hype. We identify challenges across company and stakeholder expectations, bias, explainability and overconfidence, data literacy, model lifecycles, and privacy that lead to a sobering picture: At this point in time, opinion was mixed across our organizations on whether implementing ML was even a worthwhile exercise. Through our discussion, we highlight how and why implementing this promising technology can be especially fraught for resource-constrained organizations, and so draw attention to areas requiring further study from the broader machine learning community. 






%

\begin{acks}
Thanks to Betsy Peters for providing the impetus for this work, to our respective partners for proofreading and running commentary throughout the process, and to members of the MIT Visualization Group and the MIT Interactive Robotics Group for their feedback. SB is funded by an NSF GRFP. AH is funded by a Siebel Fellowship.
\end{acks}

\bibliographystyle{ACM-Reference-Format}
\bibliography{bibliography}


\end{document}


\fancyhead{}

\title{Supplementary Material for:\\\emph{Machine Learning Practices Outside Big Tech:\\How Resource Constraints Challenge Responsible Development}}

\author{Aspen Hopkins}
\authornote{Both authors contributed equally to this research.}
\affiliation{%
  \institution{Massachusetts Institute of Technology}
  \city{Cambridge}
  \country{USA}}
\email{dataspen@mit.edu}

\author{Serena Booth}
\authornotemark[1]
\affiliation{%
  \institution{Massachusetts Institute of Technology}
  \city{Cambridge}
  \country{USA}}
\email{sbooth@mit.edu}

\renewcommand{\shortauthors}{Hopkins and Booth}




\maketitle

\subsection{Overview}
In this supplementary material, we present the outlines we used for conducting each interview. We note that we did not follow these protocols exactly as specified: we instead took a semi-structured approach to these interviews, where we followed interesting leads presented by the interviewees. We only used these protocols as rough interview guidelines.

\section{Interview with \One}
\begin{itemize}
    \item You mentioned you’re on the ML team, but you also have a separate data science team. How does that breakdown work? 
    \item What are you currently using Machine Learning for at your company? What kind of data, and what kind of models?  
    \item Going forward, what are your ML/Data aspirations? 
    \item What unrealized promise do you see in ML?
    \item What are the potential limitations? 
    \item When you or your team gets started on a data/ML project, what is your process?
    \item How do you manage these projects over time? What do updates and changes look like? 
    \item What is the life cycle of a model like at your company?
    \item What is your process for launching or sharing a data-related project with the world?
    \item How do you or would you test your models?
    \item How do you think about model regressions? How costly are failures?
    \item How do you think about bias? 
    \item Have you ever had a setback after launching a data project? What caused the setback?
    \item How do you ensure the progress you make aligns with the goals of the project and the company?
    \item What stakeholders do you work with? 
    \item What impact does communication have on you or your team?
    \item Have you experienced a situation where your team learned that some analysis was incorrect?
    \item What was the situation? How did the team handle that?
    \item How do you think your work would be different if you were at FAANG?
    \item In your regular work, what are the biggest difficulties you face regularly?
    \item ML can produce very high accuracy models, but it can be hard to understand how these models make decisions. What are your thoughts on the black box nature of these decision-making modeling tools?
    \item Has GDPR or other privacy legislation noticeably impacted or changed your work? How so?
    \item As you move forward with data-intensive projects, what are your thoughts on compliance and how easy or hard it is to be in compliance?
\end{itemize}

\section{Interview with \Two}

\begin{itemize}
    \item How are you using Machine Learning or Big Data or other automated decision tools at your company?
    \item What are the benefits of using ML or automated decision making tools? 
    \item Where are you using automated processing or tools, or ML products right now? In the future?
    \item What are the challenges you’ve faced using ML in your business or automated decision making tools? 
    \item ML can produce very high accuracy models, but it can be hard to understand how these models make decisions. What are your thoughts on the black box nature of these decision-making modeling tools?
    \item What concerns, if any, do you have about using black box systems?
    \item Have you worked to introduce explanations for your ML models? How so?
    \item What are best case scenarios from these tools? Worst case?
    \item How do you test your models? 
    \item When you or your team gets started on a data project, what is your process?
    \item What is your process for launching or sharing a data-related project with the world?
    \item How do you track these projects over time? What do updates and changes look like? 
    \item Have you ever had a setback after launching a data project? What caused the setback?
    \item Have you experienced a situation where your team learned that some analysis was incorrect? How did you handle that situation?
    \item How has GDPR affected your work?
    \item As you move forward with data-intensive projects, what are your thoughts on compliance and how easy or hard it is to be in compliance?
\end{itemize}

\section{Interview with \Three}
\begin{itemize}
    \item Our impression is that you primarily build two products. We heard you are working on product recommendation--can you describe how you’re planning to use Machine Learning or Big Data or other automated decision tools right now?
    \item What are the challenges you’ve faced or anticipate in using ML in automated decision making tools? 
    \item What are your concerns about using black box systems?
    \item What are best case scenarios from these tools? Worst case?
    \item How do you test your models? 
    \item How do you think about model regressions? How costly are failures? How do you insulate yourself from these?
    \item You’re still mostly curating recommendations manually. What goes into that decision making, and how will it change as you increasingly automate?
    \item How do you assess the quality of recommendations?
    \item What are your concerns about automating?
    \item By highlighting companies on your product you increase their reach. How do you choose which companies to highlight?
    \item When you or your team gets started on a data project, what is your process?
    \item What is your process for launching or sharing a data-related project with the world?
    \item How do you track these projects over time? What do updates and changes look like? 
    \item Which stakeholders do you share your work with? How do you go about sharing data findings with other team members?
    \item Have you ever had a setback after launching a data project?
    \item What caused the setback?
    \item Have you experienced a situation where your team learned that some analysis was incorrect? What happened? How did the team handle that?
    \item How do you think about GDPR or other privacy legislation? 
    \item As you move forward with data-intensive projects, what are your thoughts on compliance and how easy or hard it is to be in compliance?
\end{itemize}

\section{Interview with \Four}
\begin{itemize}
    \item Are you currently using Machine Learning or other automated decision tools? 
    \item What are your ML/Data aspirations? What promise do you see in ML? 
    \item How do you or would you test your models?
    \item How do you think about model regressions? How costly are failures?  
    \item When you or your team gets started on a data project, what is your process?
    \item How do you validate and assess your work?
    \item Would you ever consider external QC for data or models? 
    \item What is your process for launching or sharing a data-related project with the world?
    \item How do you track these projects over time? What do updates and changes look like? 
    \item Have you ever had a setback after launching a data project? What caused the setback?
    \item ML can produce very high accuracy models, but it can be hard to understand how these models make decisions. \item What are your thoughts on the black box nature of these decision-making modeling tools?
    \item How do you think about GDPR or other privacy legislation? 
    \item As you move forward with data-intensive projects, what are your thoughts on compliance and how easy or hard it is to be in compliance?

\end{itemize}

\section{Interview with \Five}
\begin{itemize}
    \item We had a look through your website, and we took away that you’re doing medical diagnostic support with AI. We have some questions about data, consensus-building, and model validation. 
    \item How do you collect your data?
    \item Where does it come from? 
    \item How do you label it? (How do you build consensus?)  
    \item How do you ensure sufficient representation in your data?
    \item How do you test/validate your models? 
    \item How do you track these their models and data evolution over time? 
    \item What do updates and changes look like? 
    \item How do you decide when/where to iterate on your data?
    \item Have you ever experienced a model or data regression/failures? How did you handle it? 
    \item What are the risks associated with those failures?
    \item How do you build calibrated (appropriate) trust with practitioners? 
    \item Are you worried about overtrust?
    \item How do you communicate or think about communicating uncertainty in your ML/data products? 
    \item ML can produce very high accuracy models, but it can be hard to understand how these models make decisions. What are your thoughts on the black box nature of these decision-making modeling tools?
    \item How do you think about GDPR or other privacy legislation? 
    \item As you move forward with data-intensive projects, what are your thoughts on compliance and how easy or hard it is to be in compliance?
\end{itemize}

\section{Interview with \Six}
\begin{itemize}
    \item What are you currently using Machine Learning for at your company? 
    \item What kind of data, and what kind of models?  
    \item Going forward, what are your ML/Data aspirations? \item What unrealized promise do you see in ML?
    \item What are the potential limitations? 
    \item When you or your team gets started on a data/ML project, what is your process?
    \item How do you manage these projects over time? What do updates and changes look like? 
    \item What is the life cycle of a model like at your company?
    \item What is your process for launching or sharing a data-related project with the world?
    \item How do you or would you test your models?
    \item How do you think about model regressions? How costly are failures?
    \item How do you think about bias? 
    \item How do you think about representative data/testing?
    \item Have you ever had a setback after launching a data project? What caused the setback?
    \item How do you ensure the progress you make aligns with the goals of the project and the company?
    \item What stakeholders do you work with? What impact does communication have on you or your team?
    \item Have you experienced a situation where your team learned that some analysis was incorrect?
    \item What was the situation? How did the team handle that?
    \item How do you think your work would be different if you were at FAANG?
    \item In your regular work, what are the biggest difficulties you face regularly?
    \item ML can produce very high accuracy models, but it can be hard to understand how these models make decisions. What are your thoughts on the black box nature of these decision-making modeling tools?  
    \item Has GDPR or other privacy legislation noticeably impacted or changed your work? How so?

\end{itemize}

\section{Interview with \Seven}
\begin{itemize}
    \item First let’s talk about data-related projects, not necessarily machine learning. When you or your team gets started on a data project, what is your process?
    \item How do you validate and assess your work?
    \item What is your process for launching or sharing a data-related project with the world?
    \item How do you track these projects over time? What do updates and changes look like? 
    \item Have you ever had a setback after launching a data project? What caused the setback?
    \item Have you experienced a situation where your team learned that some analysis was incorrect?
    \item What was the situation? How did the team handle that?
    \item Are you currently using Machine Learning or other automated decision tools at your company? 
    \item What are your ML/Data aspirations? What promise do you see in ML? 
    \item How do you or would you test your models?
    \item How do you think about model regressions? How costly are failures?  
    \item ML can produce very high accuracy models, but it can be hard to understand how these models make decisions. What are your thoughts on the black box nature of these decision-making modeling tools?
    \item How do you think about GDPR? 
    \item As you move forward with data-intensive projects, what are your thoughts on compliance and how easy or hard it is to be in compliance?
\end{itemize}

\section{Interview with \Eight}
\begin{itemize}
    \item Are you currently using Machine Learning or other automated decision tools? 
    \item What are your ML/Data aspirations? What promise do you see in ML? 
    \item How do you or would you test your models?
    \item How do you think about model regressions? How costly are failures?  
    \item When you or your team gets started on a data project, what is your process?
    \item How do you validate and assess your work?
    \item Would you ever consider external QC for data or models? 
    \item What is your process for launching or sharing a data-related project with the world?
    \item How do you track these projects over time? What do updates and changes look like? 
    \item Have you ever had a setback after launching a data project? What caused the setback?
    \item ML can produce very high accuracy models, but it can be hard to understand how these models make decisions. What are your thoughts on the black box nature of these decision-making modeling tools?
    \item How do you think about GDPR or other privacy legislation? As you move forward with data-intensive projects, what are your thoughts on compliance and how easy or hard it is to be in compliance?
\end{itemize}

\section{Interview with \Nine}
\begin{itemize}
    \item Are you currently using Machine Learning or other automated decision tools at your company? 
    \item What are your ML/Data aspirations? What promise do you see in ML? 
    \item How do you or would you test your models?
    \item How do you think about model regressions? How costly are failures?  
    \item When you or your team gets started on a data/ML project, what is your process?
    \item How do you validate and assess your work?
    \item What is your process for launching or sharing a data-related project with the world?
    \item How do you track these projects over time? What do updates and changes look like? 
    \item Have you ever had a setback after launching a data project? What caused the setback?
    \item Have you experienced a situation where your team learned that some analysis was incorrect?
    \item How did you find out?
    \item What was the situation? How did the team handle that?
    \item ML can produce very high accuracy models, but it can be hard to understand how these models make decisions. What are your thoughts on the black box nature of these decision-making modeling tools?
    \item How do you think about GDPR? As you move forward with data-intensive projects, what are your thoughts on compliance and how easy or hard it is to be in compliance?
\end{itemize}

\section{Interview with \Ten}
\begin{itemize}
    \item In some ways it seems like your company acts to consult companies on ML. What are the most common issues companies face?
    \item How do you work with clients? What has your experience been like?
    \item What has your experience of using ML and AI in the real world been? 
    \item What works well, and what doesn't? 
    \item What promise do you see in ML? 
    \item How do you or your clients test your models?
    \item How do you think about model regressions? How costly are failures?  
    \item Your company has a framework for data readiness. Can you expand on what that means?
    \item ML can produce very high accuracy models, but it can be hard to understand how these models make decisions. What are your thoughts on the black box nature of these decision-making modeling tools?
    \item Have you looked into the explanation/interpretability literature? 
    \item When you or your team gets started on a data project, what is your process?
    \item How do you validate and assess your work?
    \item Would you ever consider external QC for data or models? 
    \item What is your process for launching or sharing a data-related project with the world?
    \item How do you track these projects over time? What do updates and changes look like? 
    \item Have you ever had a setback after launching a data project? What caused the setback?
    \item Has your work been impacted by GDPR or other data privacy legislation?

\end{itemize}

\section{Interview with \Eleven}
\begin{itemize}
    \item We read your blurb online, but can you tell us a little about your work?
    \item How do you evaluate whether your augmentations are useful?
    \item How do you test/evaluate your models? 
    \item Related: How do you assess the quality of augmentations?
    \item What are best case scenarios from these tools? Worst case?
    \item Do you have any concerns about using black box systems?
    \item Do you ever collect new real world data instead of using computational techniques? When and why?
    \item How do you think about dataset diversity?
    \item Do you have systems in place to choose which variability to introduce? 
    \item How do you evaluate differences between testing in synthesized data vs real world contexts?
    \item If a dataset has insufficient diversity, is your approach able to compensate?
    \item Augmented data is limited by the contextual understanding of practitioners around data needs--how do you introduce real world variation?
    \item How costly are regressions/failures? What do they look like, and how do you insulate yourself?
    \item How do you track these changes to models / data?
    \item Which stakeholders do you share your work with?
    \item Do you need to educate stakeholders and customers? How do you go about this?
    \item Do you need to explain your augmentations to stakeholders/customers? How do you go about this? 
    \item Have you experienced a situation where your team learned that some model/augmentations/analysis was incorrect?
    \item Broadly, what happened was the situation? 
    \item How did the team handle that?
    \item Have you ever had a setback after launching a project?
    \item What caused the setback?
    \item Has your work been impacted by GDPR or other data privacy legislation?
\end{itemize}

\section{Interview with \Twelve}

\begin{itemize}
    \item Right now, you’re on the board of an AI startup.  Can you tell us a little bit about how you operate in this role?
    \item As a board member, what is your current involvement like? Are you involved in the company's decision making? How so? 
    \item Are you familiar with their technical work or do you focus on their business dev?
    \item What are the main challenges that they face right now?
    \item From your perspective, what does a technical failure look like for this company?
    \item What are the potential impacts of those failures? 
    \item How do you think about investing in AI/ML companies? What do you look for, and what are technological red flags for you?
    \item When AI startups need large quantities of data to make their products successful, how do you help them get off the ground? (Is this something you see)?  
    \item Do you think about the implications of GDPR-like legislation when investing in AI companies? Do you think this will change in the future, as these privacy-preserving laws become stronger and more common? 
    \item A little retrospective: What was your experience as a founder in a recommendation engine? What were the main challenges you faced?
    \item The potential of ML was just surfacing when you worked on that company. How did this play into your decision making?
    \item Has your work been impacted by GDPR or other data privacy legislation?

\end{itemize}

\section{Interview with \Thirteen}
\begin{itemize}
    \item Your company changed directions quite dramatically. How did you pivot to where you are now?
    \item How are you using Machine Learning in your product? 
    \item What are the benefits of using ML or automated decision making tools? 
    \item What are the challenges you’ve faced using ML in your business or automated decision making tools? 
    \item What are your concerns about using black box systems?
    \item What are best case scenarios from these tools? Worst case?
    \item How do you test your models? 
    \item How costly are regressions/failures? What do they look like, and how do you insulate yourself?
    \item How do you track these changes to models / data?
    \item How do you balance accuracy with generalizability?
    \item Do you have systems in place to choose which variability to introduce? How do you evaluate differences between testing in controlled contexts vs real world contexts?
    \item How does machine training translate to human interaction?
    \item What’s the user sentiment like re: ML, and how do you manage user expectations?
    \item What does onboarding look like?
    \item Do you need to explain ML decisions? How do you do this?
    \item When you or your team gets started on a data project, what is your process?
    \item What is your process for launching or sharing a data-related project with the world?
    \item How do you track these projects over time? What do updates and changes look like? 
    \item Which stakeholders do you share your work with? 
    \item How do you go about sharing data findings with other team members?
    \item Have you ever had a setback after launching a data project? What caused the setback?
    \item Have you experienced a situation where your team learned that some analysis was incorrect? Broadly, what happened was the situation? 
    \item ML can produce very high accuracy models, but it can be hard to understand how these models make decisions. What are your thoughts on the black box nature of these decision-making modeling tools?
    \item Has your work been impacted by GDPR or other data privacy legislation?

\end{itemize}

\section{Interview with \Fourteen}

\begin{itemize}
    \item Outside of data management, what are your office’s responsibilities? How much do you collaborate with other agencies and organizations?
    \item When you collect data, why do you collect data? 
    \item Where/what kinds of data are important?
    \item Obviously there’s a cost to introducing a new dataset. What do you want more of?
    \item Have you made mistakes by introducing datasets before?
    \item How do you collect your data? (How much is internally completed vs relying on outside observers?)
    \item How do you evaluate data collected by other parties?
    \item In general, what quality controls do you use when evaluating data? (what is the quality assurance system?) 
    \item How do you ensure sufficient representation in your data?
    \item How do you represent marginalized communities/mitigate bias in your data?
    \item How do you uncover discrepancies between data collected and real world distributions?
    \item How do you handle data redundancies? 
    \item How do you label it (if you label it)? 
    \item How do you track data changes over time? 
    \item You mentioned refining data processes for 311 team following COVID-19 tracking and response. What were the specific changes you made and why? How did you implement those?
    \item What are the major difficulties you face when communicating around or about data to other groups?
    \item How does data literacy play into these dynamics? 
    \item You mentioned one of your major focuses is on building buy in/trust in data or analytics for other orgs. What specific difficulties do you face in doing so?
    \item Do you/your team use your datasets to create models? If so, could you tell us a bit about these?
    \item How do you test these models?
    \item Do you fact check other models?
    \item What are your thoughts on data privacy? Are you concerned about collecting too much data?
    \item How has GDPR affected your work?
    \item There’s a trend of using data to mislead the public---intentionally, or not. How do you think about misinformation, and the role of open data?
    \item What does it mean for data to be accessible?

\end{itemize}
\section{Interview with \Fifteen}

\begin{itemize}
    \item How do you think about investing in machine learning?
   \item Are you using Machine Learning or Big Data or other automated decision tools at your company?
    \item How so? 
    \item If not, why not? 
    \item What are the benefits of using ML or automated decision making tools? 
    \item Where are you using automated processing or tools, or ML products right now? In the future?
    \item What are the challenges you’ve faced using ML in your business or automated decision making tools? 
    \item ML can produce very high accuracy models, but it can be hard to understand how these models make decisions. What are your thoughts on the black box nature of these decision-making modeling tools?
    \item What concerns, if any, do you have about using black box systems?
    \item What are best case scenarios from these tools? Worst case?
    \item How do you test your models? 
    \item When you or your team gets started on a data project, what is your process?
    \item What is your process for launching or sharing a data-related project with the world?
    \item How do you track these projects over time? What do updates and changes look like? 
    \item Have you ever had a setback after launching a data project?
    \item What caused the setback?
    \item Have you experienced a situation where your team learned that some analysis was incorrect?
    \item Broadly, what happened was the situation? 
    \item How did the team handle that?
    \item How has GDPR affected your work?
\end{itemize}

\section{Interview with \Sixteen}
\begin{itemize}
    \item Are you currently using Machine Learning or other automated decision tools at your company? 
    \item What are your ML/Data aspirations? What promise do you see in ML? 
    \item How do you or would you test your models?
    \item How do you think about model regressions? How costly are failures?  
    \item When you or your team gets started on a data/ML project, what is your process?
    \item How do you validate and assess your work?
    \item What is your process for launching or sharing a data-related project with the world?
    \item How do you track these projects over time? What do updates and changes look like? 
    \item Have you ever had a setback after launching a data project? What caused the setback?
    \item Have you experienced a situation where your team learned that some analysis was incorrect?
    \item What was the situation? How did the team handle that?
    \item You work on language translation projects. How do you think about your users needs in this space? 
    \item ML can produce very high accuracy models, but it can be hard to understand how these models make decisions. What are your thoughts on the black box nature of these decision-making modeling tools?
    \item How do you think about GDPR? As you move forward with data-intensive projects, what are your thoughts on compliance and how easy or hard it is to be in compliance?

\end{itemize}

\section{Interview with \Seventeen}
\begin{itemize}
    \item Are you currently using Machine Learning or other automated decision tools at your company? 
    \item What are your ML/Data aspirations? What promise do you see in ML? 
    \item How do you or would you test your models?
    \item How do you think about model regressions? How costly are failures?  
    \item When you or your team gets started on a data/ML project, what is your process?
    \item How do you validate and assess your work?
    \item What is your process for launching or sharing a data-related project with the world?
    \item How do you track these projects over time? What do updates and changes look like? 
    \item Have you ever had a setback after launching a data project? What caused the setback?
    \item Have you experienced a situation where your team learned that some analysis was incorrect?
    \item What was the situation? How did the team handle that?
    \item You work on language translation projects. How do you think about your users needs in this space? 
    \item ML can produce very high accuracy models, but it can be hard to understand how these models make decisions. What are your thoughts on the black box nature of these decision-making modeling tools?
    \item How do you think about GDPR? As you move forward with data-intensive projects, what are your thoughts on compliance and how easy or hard it is to be in compliance?

\end{itemize}
